# Improving Low-Light Image Recognition Performance Based on Image-adaptive Learnable Module

Seitaro Ono[1], Yuka Ogino[2], Takahiro Toizumi[2], Atsushi Ito[2] and Masato Tsukada[1,2]
[1]*University of Tsukuba, Ibaragi, Japan*
[2]*NEC Corporation, Kanagawa, Japan*
*seitaro.ono@image.iit.tsukuba.ac.jp, tsukada@iit.tsukuba.ac.jp, {yogino, t-toizumi_ct, ito-atsushi}@nec.com,*

Keywords: Low-Light Image Enhancement, Image Recognition

Abstract: In recent years, significant progress has been made in image recognition technology based on deep neural networks. However, improving recognition performance under low-light conditions remains a significant challenge. This study addresses the enhancement of recognition model performance in low-light conditions. We propose an image-adaptive learnable module which apply appropriate image processing on input images and a hyperparameter predictor to forecast optimal parameters used in the module. Our proposed approach allows for the enhancement of recognition performance under low-light conditions by easily integrating as a front-end filter without the need to retrain existing recognition models designed for low-light conditions. Through experiments, our proposed method demonstrates its contribution to enhancing image recognition performance under low-light conditions.

## 1 INTRODUCTION

In recent years, image recognition with deep neural networks (DNNs) has advanced significantly. Various recognition models, trained on large datasets, have emerged and improved steadily in performance. These models predominantly assume inputs of high-quality images captured in well-lighting conditions. The challenge of image recognition still remains in adapting to various real-world conditions. In practical applications, environmental factors such as low lighting, backlighting, adverse weather, and image sensor noise significantly impact image quality. These factors degrade the performance of image recognition.

To overcome these issues, many learning-based image enhancement methods have been proposed so far. These methods primarily aim to improve perceptibility in human vision and do not necessarily focus on enhancing the performance of recognition models. In particular, learning-based low-light image enhancement methods are designed without considering the subsequent recognition task. These methods may overly smooth images or accentuate noise in images, leading to a decline in recognition performance.

Recently, a method has been proposed to enhance recognition performance under extremely low-light conditions. Lee et al. proposed a human pose estimation model which estimates poses of individuals in extremely low-light images (Lee et al., 2023). The pose estimation model is well trained on images captured in extremely low-light conditions. The model has significantly contributed to improving the accuracy of pose estimation. However, we hypothesize that further enhancement of performance is achievable by introducing crucial image quality enhancement for the model.

We propose an image-adaptive learnable module and a hyper-parameter predictor to optimally process input images to improve the performance of the later stage recognition task. The proposed method does not aim to improve image quality in human perception. Instead, it focuses on enhancing recognition model performance through the enrichment of useful features for the model. We introduce a relatively straightforward image processing module to correct low-light images into images that can be easily recognized by downstream recognition models. Furthermore, to enhance recognition performance, we propose a method for predicting appropriate hyperparameters within the image processing module.

In this paper, we adopt Lee et al.'s pose estimation as the recognition model and further improve the performance of the model. Through experiments, we demonstrate that our proposed approach enhances the recognition performance even with recognition models trained on low-light images beforehand, validating its practical effectiveness. This research introduces a new direction for image recognition in real-world environments, providing a crucial foundation for advancing the performance of existing recognition models.

## 2 RELATED WORKS

Research on improving the image quality of low-light images is considered an important challenge in the fields of computer vision and image processing. Various techniques have been proposed to enhance the lightness of low-light images and improve their overall quality.

In recent years, with the rapid development of deep learning, correction techniques for low-light images using Convolutional Neural Networks (CNNs) have gained attention. These methods can learn features from low-level image characteristics to high-level semantic features and improve image quality in low-light environments. These approaches not only increase the lightness of an image but also achieve advanced image restoration, such as color correction and noise reduction. LLNet (Lore et al., 2017) employs a deep autoencoder for enhancing low-light images and noise reduction. It enables end-to-end training by appropriately adjusting lightness while preserving the natural appearance of the image. MSRNet (Shen et al., 2017) learns the mapping between dark and bright images by using different Gaussian convolution kernels in an end-to-end manner. MBLLEN (Lv et al., 2018) utilizes a multi-branch network to extract rich features at different levels and ultimately generates the output image through multi-branch fusion. RetinexNet (Wei et al., 2018) combines the Retinex theory (Land et al. 1977) with CNNs to estimate the illumination map of an image. It improves low-light images by adjusting the map. KinD (Zhang et al., 2019) is a network based on the Retinex theory, designed with an additional Restoration-Net for noise removal. This approach effectively adjusts the illumination of low-light images and achieves high-quality image reconstruction by reducing noise.

These methods are designed to address critical challenges in low-light image processing, such as realistic image reconstruction, noise removal, and management of lightness and contrast. However, many conventional data-driven methods heavily rely on large sets of paired data of dark and bright images. The cost of collecting such paired datasets has increased, imposing practical constraints.

These conventional methods primarily focus on improving visibility for human perception, without considering the subsequent recognition tasks. Therefore, when applying these methods, there is a concern that crucial features for recognition tasks may be lost during image processing. In other words, there is a concern about whether it is appropriate to directly apply these methods to recognition tasks. Thus, it is difficult to strike a balance between improving the quality of the input image and the accuracy of the recognition task.

Image-Adaptive YOLO (Liu et al., 2022) addresses this issue by jointly training an image processing module along with the subsequent object detection model. The image processing handles degraded input images captured under adverse weather or low-light conditions. This approach strikes a balance between enhancing input image quality and improving the accuracy of the object detection model. It improves object detection accuracy in adverse weather and low-light conditions.

However, the cooperative learning approach between the frontend image processing module and the subsequent recognition model is not necessarily the optimal solution. As the subsequent recognition model becomes large-scale, training from scratch demands a computationally expensive environment and extensive processing time. In the current situation where various recognition models are continually proposed, the situation of incurring high costs and extensive processing time with each new model training from scratch is not desirable. There is a demand for the development of recognition-model-centric (low-light) image enhancement methods that offer outstanding efficiency and flexibility, eliminating the necessity to train the subsequent recognition model.

Lee et al. (Lee et al., 2023) proposed a method for estimating the poses of individuals in images captured under extremely dark lighting conditions. They developed a camera that can simultaneously capture dark and well-exposed images of the same scene. By controlling the intensity of light, the camera can simultaneously capture both images of a scene under a dark environment and a bright environment. Their pose estimation model integrates a model processing images under low-light conditions with a model processing corresponding high-exposure images. It learns representations independent of lighting conditions and improves individual pose estimation accuracy under extremely low-light conditions based on these representations. Because this method does not perform image enhancement on the input dark images, however, it is not possible to check the consistency of the model's prediction results against the actual captured images.

# 3 PROPOSED METHOD

In recognition applications dealing with images or videos captured in low-light environments, a significant challenge is the degradation of recognition accuracy. When a very dark image taken in a low-light environment is input, it becomes difficult to distinguish between the subject and the background because of the low contrast between the subject and the background. Therefore, in various image recognition tasks, the extraction of feature values to identify the subject is hindered. This difficulty in extracting feature values of the subject impedes proper identification. Additionally, a contributing factor to the difficulty of recognition tasks in low-light environments is the noise originating from image sensors. In dark images captured in low-light conditions, high levels of noise occur, obscuring the fundamental structure of the scene. Differentiating crucial features values from random noise becomes challenging, leading to incorrect recognition results.

To overcome this problem, this study proposes a Low-Light Enhancement (LLE) framework that adapts input images by recovering exposure and removing noise, making it easier to extract potential feature values crucial for downstream recognition tasks.

As mentioned earlier, Lee et al. proposed a method for accurately estimating human poses from images under extremely low-light conditions. In the method, the pose estimation model learns the similarity of feature representations between appropriately exposed images and images captured under extremely low-light conditions. However, the method does not actively perform contrast adjustment or noise reduction, thereby inadequately addressing the degradation components of low-light images. Consequently, these aspects may limit the pose estimation model from fully unleashing its latent performance.

We introduce the Low-Light Enhancement (LLE) framework as a front-end module. This framework explicitly incorporates mechanisms for exposure recovery, contrast adjustment, and noise reduction. By doing so, it mitigates challenges posed by low-light conditions and facilitates the extraction of crucial features for downstream recognition tasks.

Unlike Image-Adaptive YOLO, our proposed approach focuses solely on training the LLE part independently of the downstream recognition model. This approach allows training only the LLE part without modifying the pre-trained recognition model, achieving performance improvement as a front-end filter. This feature makes it easily applicable to various existing pre-trained recognition models in future. The proposed method not only recovers exposure and removes noise from input images but also enhances them to facilitate the extraction of feature values tailored for downstream recognition tasks. Figure 1 (a) illustrates the framework. The entire pipeline comprises a differentiable image processing module consisting of differentiable multiple image processing operators, a Fully Convolutional Network (FCN)-based optimal parameter predictor to predict optimal parameters (LLE parameters) for image processing operators, and a recognition model. Initially, an input image is randomly cropped to a size of 256×256 and fed into the optimal parameter predictor to predict LLE parameters for the differentiable image processing modules. The optimal parameter predictor undergoes end-to-end training, considering recognition loss to calculate LLE parameters that maximize the recognition performance of the recognition model. The differentiable image processing module applies the LLE parameters obtained from the optimal parameter predictor to image processing operators. The image processing operators process the entire original image with the LLE parameters and generates an input image for the recognition model.

## 3.1 Differentiable Image Processing Module

To enable gradient-based optimization for the optimal parameter predictor, all the various image processing operators used within the differentiable image processing module need to be differentiable. Our proposed differentiable image processing module consists of three differentiable image processing operators with adjustable hyperparameters: Exposure, Gamma, and Smoothing (Denoising). Among these,

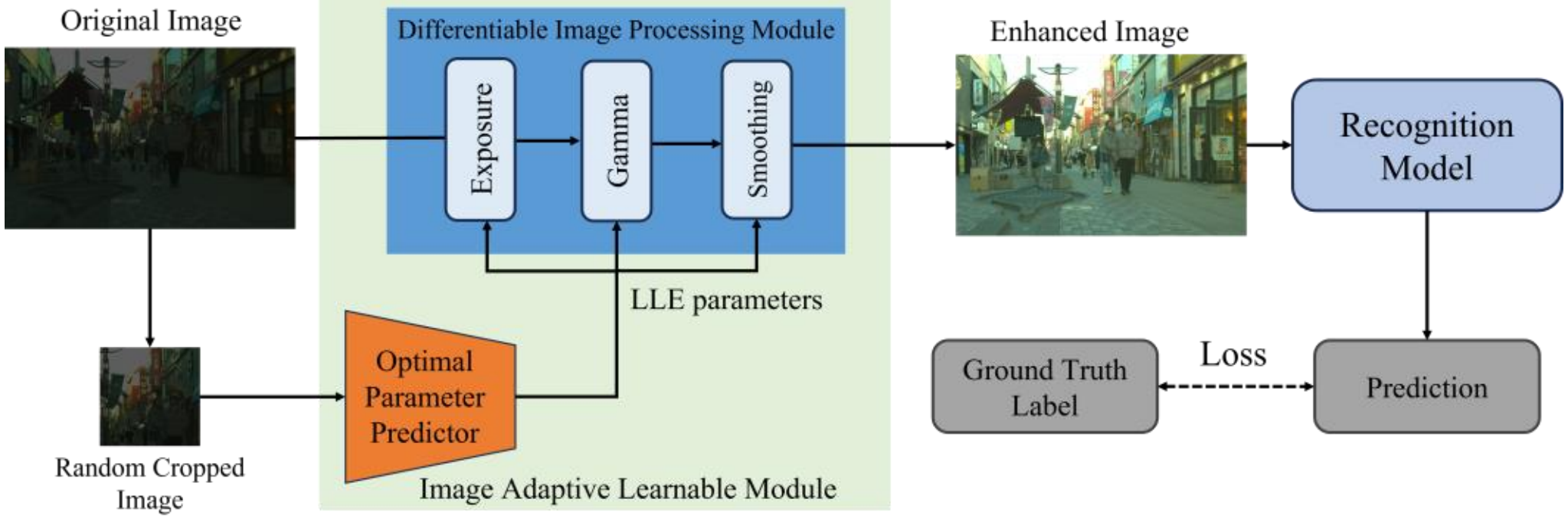

(a) The overall framework of the proposed Low-Light Image Enhancement method.

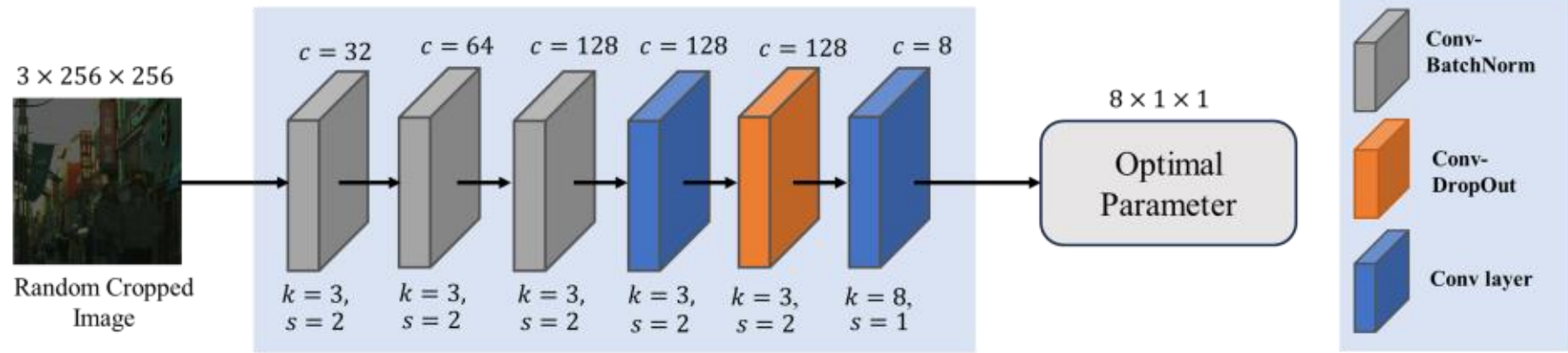

(b) The architecture of the proposed optimal parameter predictor.

Figure 1: (a) End-to-end training pipeline of the proposed low-light image enhancement method. The optimal parameter predictor predicts the best parameters (LLE parameters) for the differentiable image processing module from randomly cropped images. The original images are processed by the differentiable image processing module, enhancing the performance of the subsequent recognition model. During training, the recognition model does not undergo gradient updates. The parameter predictor learns to minimize task-specific losses between the predictions of the recognition model and ground truth data. (b) Configuration of the proposed parameter predictor. The predictor is a Fully Convolutional Network (FCN) consisting of six convolutional layers. "k", "s" and "c" respectively denote the kernel size, the stride and the number of output channels for each convolutional layer.

Exposure and Gamma operators perform pixel-wise arithmetic operations. The Smoothing (Denoising) operator is specifically designed to suppress noise components without losing content information in the image.

#### 3.1.1 Exposure Operator

This operator adjusts the overall lightness of the entire image by raising or lowering the exposure level, effectively controlling the overall lightness. For an input pixel value $P_i = (r_i, g_i, b_i)$ and an output pixel $P_o = (r_o, g_o, b_o)$, the operator performs the following mapping:

$$P_o = aP_i, \tag{1}$$

where, $a$ is the parameter predicted by the parameter predictor.

#### 3.1.2 Gamma Operator

The Gamma operator alters the contrast of the image, emphasizing or de-emphasizing specific details by changing the gamma value. For an input pixel value $P_i = (r_i, g_i, b_i)$ and an output pixel $P_o = (r_o, g_o, b_o)$, the operator performs the following mapping:

$$P_o = P_i^{\gamma}, \tag{2}$$

where, $\gamma$ is the parameter predicted by the parameter predictor. The operations in the Exposure operator and Gamma operator involve simple multiplication and exponentiation, making them differentiable.

#### 3.1.3 Smoothing Operator

The Smoothing operator smoothens the input image while preserving edge information crucial for the recognition model. To achieve this, we adopted a bilateral filter (Tomasi et al., 1998). For an 1-channel image with $I \times J$ pixels, let $f(i,j)$ represent input pixel value at spatial coordinates $(i,j)$, if we apply the bilateral filter to $f(i,j)$, resulting in the output pixel value $g(i,j)$, it can be expressed as follows:

$$g(i,j) = \frac{\sum_{n=-w}^{w}\sum_{m=-w}^{w} f(i+m,j+n) e^{-\frac{m^2+n^2}{2\sigma_1^2}} e^{-\frac{(f(i,j)-f(i+m,j+n))^2}{2\sigma_2^2}}}{\sum_{n=-w}^{w}\sum_{m=-w}^{w} e^{-\frac{m^2+n^2}{2\sigma_1^2}} e^{-\frac{(f(i,j)-f(i+m,j+n))^2}{2\sigma_2^2}}}, \quad (3)$$

where, $\sigma_1$ and $\sigma_2$ are parameters provided by the parameter predictor, and $W$ is the window size. $\sigma_1$ adjusts the influence of the distance between coordinates $(i,j)$ and $(i+m,j+n)$, with a larger $\sigma_1$ reducing the impact of pixels that are farther away. $\sigma_2$ adjusts the influence of the difference between $f(i,j)$ and $f(i+m,j+n)$, with a larger $\sigma_2$ reducing the impact of pixels with a larger difference in values. We apply this operator independently to the three channels of RGB.

## 3.2 Optimal Parameter Predictor

In the camera's Image Signal Processing (ISP) pipeline, various image processing operators are performed to obtain images with high visibility for human perception. The correction parameters for each of image processing operators are traditionally determined empirically by experienced technicians. (Mosleh et al., 2020). The tuning process to obtain correction parameters suitable for a diverse range of images incurs substantial costs. Moreover, in this study, the objective is not to improve human perceptibility but to seek correction parameters that maximize the recognition performance of the downstream recognition model.

To address this issue, we employ an efficient small Fully Convolutional Network (FCN) as the parameter predictor to estimate LLE parameters for each input image. The purpose of the optimal parameter predictor is to understand aspects such as the exposure level and noise level in the input image and predict the LLE parameters for the image processing operators that maximize the recognition performance of the downstream recognition. Since FCNs consume considerable computational resources when processing high-resolution images, the LLE parameters are learned for randomly cropped images of 256 pixels × 256 pixels from the input image.

In real-world environmental scenes, illumination intensity is not necessarily constant, and the exposure level and noise level are not globally constant. However, for the sake of computational efficiency, we prioritize the benefit of significantly reducing computational costs and use randomly cropped images as inputs to the optimal parameter predictor.

During training, the optimal parameter predictor references the recognition loss derived from the recognition results of the downstream recognition model and learns to maximize recognition accuracy.

As mentioned in the introduction, we adopt single-person pose estimation by Lee et al. as the recognition task. Therefore, we utilize the pose estimation loss widely used in pose estimation tasks as the loss function. The pose estimation loss is represented by the following formula:

$$Loss = \frac{1}{K}\sum_{i=1}^{K} \|P_i - X_i\|_2^2, \quad (4)$$

where $P_i$ and $X_i$ represent the predicted heatmap and ground truth heatmap of the $i$-th pose estimation model, respectively, and $K$ denotes the number of keypoints.

As shown in Figure 1 (b), the optimal parameter predictor consists of 6 convolutional layers. Except for the 4th, 5th and 6th layers, each convolutional layer is followed by a Batch Normalization layer, which normalizes the distribution of input data, suppresses data variability caused by changes in lighting conditions, and enables the model to make consistent predictions. Batch Normalization also stabilizes the distribution of gradients, promoting the convergence of training. A Dropout layer is applied after the 5th convolutional layer. The final layer outputs the hyperparameters for the differentiable image processing module. The parameter predictor has only 455k parameters, given a total of 8 hyperparameters for the differentiable image processing module.

# 4. EXPERIMENTS

We evaluated the performance of our method for images captured in low-light environments. Additionally, we conducted two ablation studies to investigate the performance of each proposed differentiable image processing module and the impact of the order of the image processing operators.

## 4.1 Implementation Details

In this experiment, we adopt the pose estimation model proposed by Lee et al. as the recognition model. This model is pre-trained on the low-light image dataset for pose estimation, known as the ExLPose dataset (Lee et al., 2023). The combination of image processing operators used in our differentiable image processing module is [Exposure, Gamma,

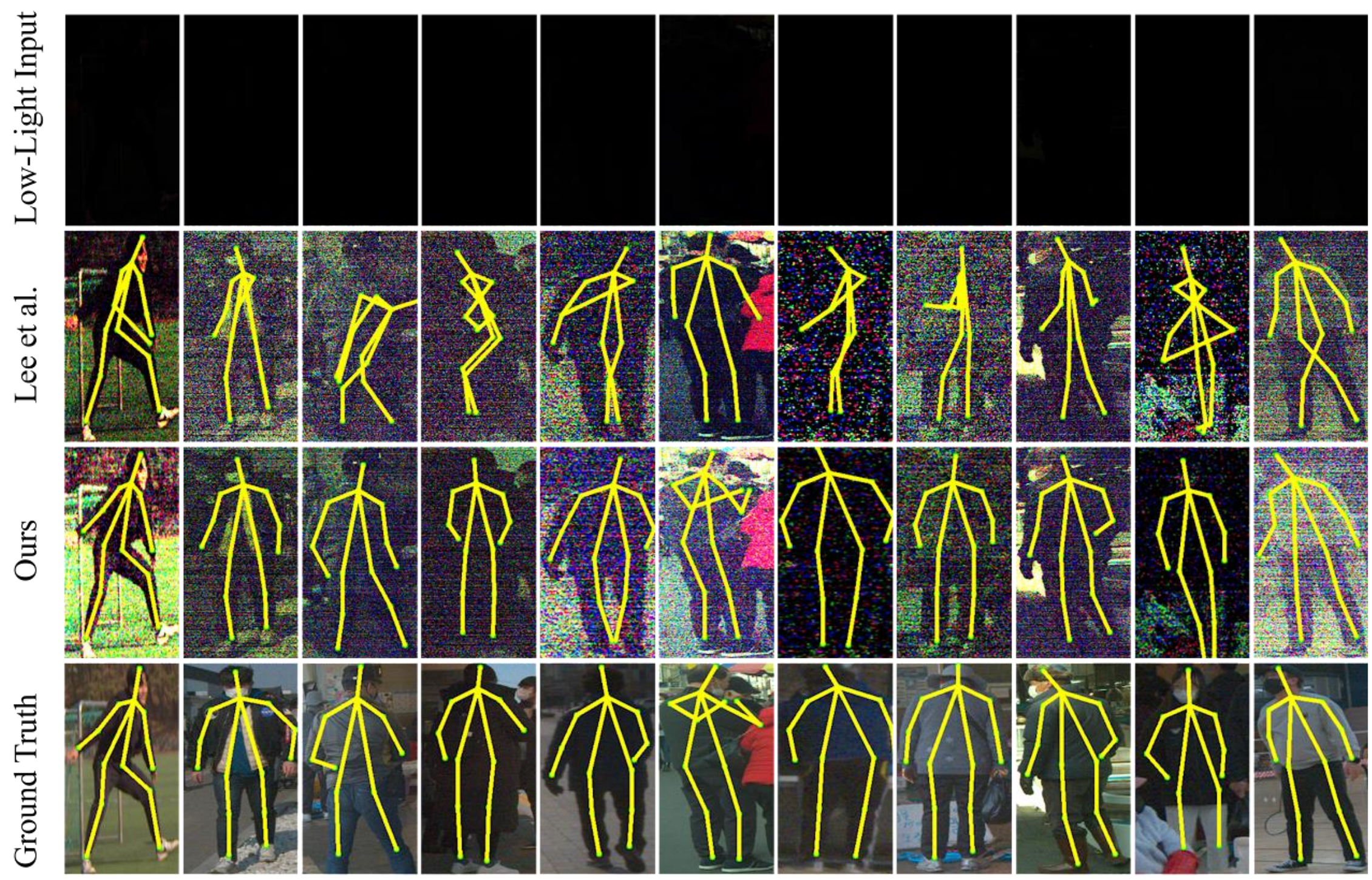

Figure 2: Qualitative evaluation results. The first row represents the input low-light image, the second row shows the image whose lightness values have been shifted to an average of 0.4 using Lee's method, the third row displays the image processed by our proposed method, and the fourth row contains the corresponding bright images paired with the input low-light images. Our proposed method effectively controls exposure and contrast in a manner that is easily understandable for the recognition model, resulting in the recovery of recognition accuracy.

Smoothing]. The optimal parameter predictor is trained to maximize the performance of Lee's pose estimation model. During the training, parameters of the pose estimation model are fixed. Only gradients of the pose estimation model are back propagated from loss function to update trainable parameters of the optimal parameter predictor. The optimal parameter predictor uses the Adam Optimizer with a learning rate set to 1e-4 and is trained for 10 epochs with a batch size of 8. We conducted the experiment using PyTorch and executed it on a GTX 1080Ti GPU.

## 4.2 Dataset

In this experiment, we used ExLPose dataset (Lee et al. 2023). ExLPose dataset is designed for human poses estimation. It has paired extremely low-light (LL) and well-exposed images captured at the same scene with the same optical axis. The light intensities of the LL images are reduced to 1/100 from those of well-exposed images by using ND filters in their camera. This dataset contains LL images, their paired well-exposed images, and the ground truth human pose labels, comprising 2065 training data pairs and 491 test data pairs. In this study, we used only the LL images for both training and testing.

Table 1: Evaluation result on ExLPose Dataset. Our proposed method improves accuracy by performing adaptive image processing on the input image.

| | AP@0.5-0.95 ↑ | | | |
|---|---|---|---|---|
| Model | LL-N | LL-H | LL-E | LL-A |
| Lee et al., 2023 | 42.1 | 33.8 | 18.0 | 32.4 |
| Ours + Lee et al., 2023 | ***42.6*** | ***34.1*** | ***20.0*** | ***33.2*** |

## 4.3 Evaluation Protocol

In this experiment, we follow the same evaluation method as Lee et al. Since we adopt single-person pose estimation as the recognition task, we assume that true bounding boxes are provided for each person in the image. The evaluation metric used is the Standard Average Precision (AP) score based on the detected Object's Keypoint Similarity (OKS), widely used in pose estimation tasks. The low-light test

Table2: Ablation analysis on the Differentiable Image Processing Module. E, G, and S stand for Exposure, Gamma, and Smoothing operators, respectively.

| | AP@0.5-0.95 ↑ | | | |
|---|---|---|---|---|
| Operator | LL-N | LL-H | LL-E | LL-A |
| E, G | 40.5 | 32.3 | 18.0 | 31.5 |
| E, S | 37.3 | 31.6 | 19.1 | 30.0 |
| G, S | 20.2 | 8.0 | 1.1 | 10.3 |
| E, G, S | ***42.6*** | ***34.1*** | ***20.0*** | ***33.2*** |

images are divided into subsets based on their respective mean lightness: LL-Normal (LL-N), LL-Hard (LL-H), and LL-Extreme (LL-E). The mean pixel intensity of LL-N, LL-H, and LL-E images are 3.2, 1.4, and 0.9, respectively, indicating that all of them can be classified as extremely dark images. The union of all three low-light subsets is denoted as LL-All (LL-A).

## 4.4 Experimental Results

The optimal parameter predictor receives randomly cropped images as input, causing variability in the recognition model's predictions for each evaluation. In this experiment, we conducted three evaluations for each of the four subsets of the ExLPose dataset (LL-N, LL-H, LL-E, LL-A), reporting the average values to account for this variability. In this experiment, we compared the accuracy of human pose estimation when our proposed method was applied and when it was not applied to Lee's pose estimation model. The comparison results of the pose estimation accuracy are shown in Table 1. Applying our proposed method to Lee's pose estimation model led to improved pose estimation accuracy in all subsets. Specifically, there was an improvement of 1.2% (0.5 points) in LL-N, 1.0% (0.3 points) in LL-H, 11.1% (2.0 points) in LL-E, and 2.5% (0.8 points) in LL-A. These results demonstrate that the proposed method can enhance significant performance improvement observed in the extremely low-light condition of LL-E. Based on these experimental results, the performance of the proposed method has been confirmed. Figure 2 shows qualitative evaluation results. Comparing the results of applying Lee et al.'s proposed input data normalization method in the second row to the application of our proposed method to the images in the third row, it is evident

Table3: Evaluation results when changing the order of the three operators. The best results were obtained when the order was E (Exposure), G (Gamma), S (Smoothing).

| | AP@0.5-0.95 ↑ | | | |
|---|---|---|---|---|
| Operator | LL-N | LL-H | LL-E | LL-A |
| S, E, G | 36.9 | 29.6 | 15.7 | 28.4 |
| S, G, E | 6.4 | 8.9 | 3.1 | 6.1 |
| G, E, S | 9.6 | 10.7 | 4.5 | 8.5 |
| G, S, E | 8.0 | 9.8 | 2.7 | 6.9 |
| E, S, G | 40.5 | 32.3 | 17.9 | 31.5 |
| E, G, S | ***42.6*** | ***34.1*** | ***20.0*** | ***33.2*** |

that our proposed method in the third row is much closer to the ground truth in the fourth row. This demonstrates the performance of our proposed approach in performing optimal image processing based on the characteristics of the pretrained recognition model in the subsequent stage.

## 4.5 Ablation Study

To validate the performance of each operator in the differentiable image processing module, we evaluated combination of the three image processing operators using four subsets of the ExLPose test dataset (LL-N, LL-H, LL-E, LL-A). Table 2 shows evaluation results of the combination of two operators. For each combination, we trained the optimal parameter predictor with the same training settings. We additionally show the result of three operators (E, G, S). The result of the three image processing operators yielded the best performance, demonstrating the effectiveness of these operators.

Furthermore, we investigated the performance of the order of the proposed three image processing operators. We swapped the order of three proposed image processing operators, trained the optimal parameter predictor with the same training settings. Table 3 shows the results of the performance of all the orderings. The results revealed that in the order of [Exposure, Gamma, Smoothing] is crucial for higher performance of the pose estimation task. Adjusting exposure spreads pixel values in low-light regions linearly. It expands the overall range of pixel values and enhances details and features. Subsequent gamma adjustment makes natural and uniform lightness distribution. It can extract detailed information from low-light images. On the other hand, the ordering that switches Exposure operator and Gamma operator significantly decreases the recognition performance. The inverse order makes nonlinear transformations to information biased towards low-light regions. It potentially destroys structural information in the image. It also makes it difficult to extract features for recognition tasks and leads to a degradation in recognition performance.

Table 4: Results of the accuracy comparison when performing image processing based on the optimal parameters obtained through grid search for each test data image.

| | AP@0.5-0.95 ↑ | | | |
|---|---|---|---|---|
| Model | LL-N | LL-H | LL-E | LL-A |
| Lee et al., 2023 | 42.1 | 33.8 | 18.0 | 32.4 |
| Ours + Lee et al., 2023 | ***51.5*** | ***41.4*** | ***27.0*** | ***40.9*** |

## 4.6 Discussion

Our approach improves accuracy without retraining the recognition model by incorporating exposure recovery and noise reduction into the image processing pipeline. This suggests that by applying our proposed method, it became possible to retrieve the overlooked features in Lee et al.'s pose estimation. To explore the potential performance of our proposed method, we conducted a grid search on the entire test data of ExLPose. In this process, we searched for optimal parameters for each input image and processed the input images in the differentiable image processing module. The processed images were then input into Lee et al.'s recognition model. The results are presented in Table 4. As evident from the results of the preliminary experiment, the pose estimation accuracy significantly improved across all subsets. This suggests the potential to further enhance the performance of the proposed method. The refinement of the training method for the optimal parameter predictor will be a future task.

# 5 CONCLUSIONS

We proposed an image-adaptive learnable module that improves recognition performance in low-light environments without retraining the pretrained recognition model for pose estimation. Our proposed method consists of a differentiable image processing module and an optimal parameter predictor. The Differentiable image processing module restores the exposure and remove noise from low-light images to recover the latent content of the images. The optimal parameter predictor predicts the optimal hyperparameters used in the modules by using a small FCN. The entire framework was trained end-to-end, and the optimal parameter predictor learned to predict appropriate hyperparameters by referring only to the loss of the subsequent pose estimation task in this paper. The experimental results demonstrated that our approach achieved a maximum recovery of up to 11.1% in the accuracy of pretrained pose estimation models across different levels of low-light image data.